\documentclass[10pt, conference, compsocconf]{IEEEtran}
\usepackage{cite}
\usepackage{amsmath,amssymb,amsfonts}
\usepackage{algorithmic}
\usepackage{graphicx}
\usepackage{textcomp}
\usepackage{xcolor}
\usepackage{caption}
\usepackage{subcaption}
\usepackage{rotating}
\usepackage{todonotes}
\usepackage{dblfloatfix}  
\usepackage{url}
\def\BibTeX{{\rm B\kern-.05em{\sc i\kern-.025em b}\kern-.08em
    T\kern-.1667em\lower.7ex\hbox{E}\kern-.125emX}}

\ifCLASSINFOpdf
\else
\fi
\begin{document}
%
\title{GalaxAI: Machine learning toolbox for interpretable analysis of\\spacecraft telemetry data
\thanks{\dag - Contributed equally and should be considered as joint first authors.\\
Corresponding authors: ana.kostovska@ijs.si, matej.petkovc@ijs.si, tomaz.stepisnik@ijs.si, dragi.kocev@ijs.si}
}


\author{\IEEEauthorblockN{Ana Kostovska$^{1,2,\dag,}$, Matej Petkovi\'{c}$^{1,2,\dag}$, Toma\v{z} Stepi\v{s}nik$^{1,2,\dag}$, Luke Lucas$^3$, Timothy Finn$^4$, Jos\'{e}  Mart\'{i}nez-Heras$^5$,\\ Pan\v{c}e Panov$^{1,2}$, Sa\v{s}o D\v{z}eroski$^2$, Alessandro Donati$^4$, Nikola Simidjievski$^{1,2,6}$, Dragi Kocev$^{1,2}$}
\IEEEauthorblockA{
\textit{$^1$Bias Variance Labs, Ljubljana, Slovenia} \\
\textit{$^2$Jo\v{z}ef Stefan Institute, Ljubljana, Slovenia} \\
\textit{$^3$LSE Space GmbH, Gilching, Germany} \\
\textit{$^4$ESOC, European Space Agency, Darmstadt, Germany} \\
\textit{$^5$Solenix Engineering, Darmstadt, Germany} \\
\textit{$^6$University of Cambridge, Cambridge, UK} \\
\textit{\dag - Contributed equally and should be considered as joint first authors.}\\
Corresponding authors: ana.kostovska@ijs.si, matej.petkovc@ijs.si, tomaz.stepisnik@ijs.si, dragi.kocev@ijs.si}
}

%


\maketitle

\begin{abstract}
We present GalaxAI - a versatile machine learning toolbox for efficient and interpretable end-to-end analysis of spacecraft telemetry data. GalaxAI employs various machine learning algorithms for multivariate time series analyses, classification, regression, and structured output prediction, capable of handling high-throughput heterogeneous data. These methods allow for the construction of robust and accurate predictive models, that are in turn applied to different tasks of spacecraft monitoring and operations planning. More importantly, besides the accurate building of models, GalaxAI implements a visualisation layer, providing mission specialists and operators with a full, detailed, and interpretable view of the data analysis process. We show the utility and versatility of GalaxAI on two use-cases concerning two different spacecraft: i) analysis and planning of Mars Express thermal power consumption and ii) predicting of INTEGRAL's crossings through Van Allen belts.
\end{abstract}


%
\IEEEpeerreviewmaketitle

\section{Introduction}

Spacecraft operate in extremely challenging and unforgiving environments. This calls for careful planning of their operations and close monitoring of their status and health \cite{McGovern2011}. The spacecraft's monitoring includes analysing housekeeping telemetry data that measure and describe the spacecraft's status, its activities, and its environment. These include temperature values at different locations, radiation values, power consumption estimates, status/command execution of active onboard equipment, performed computational activities \cite{Yairi17:jrnl,Carlton2018:jrnl,Ahn2020:jrnl,Wang2019:jrnl,heras2014:jrnl,Ibrahim19:jrnl,pilastre2020}. 

Analysing telemetry data is complex and nontrivial since typically such data is: high dimensional (the number of features easily reaches the thousands); multimodal (data measured from different onboard components at different times); heterogeneous (the variables describing the status can be of different data types); with temporal dependence (the housekeeping data are typically multidimensional time series); has missing values (different sampling periods and timings, not always all values of all variables are retrieved); and contains obvious outliers (extreme abnormal values caused by errors in data conversion or transmission) \cite{Yairi17:jrnl}. 

Based on the analysis of these telemetry data, the spacecraft mission-planning and operations teams make decisions about the spacecraft's next operations - what activities it will perform (in terms of its mission) and when it will perform them. In this paper, we present a machine learning toolbox for efficient and interpretable end-to-end data analysis of spacecraft telemetry data. We showcase its potential by analysing telemetry data of two spacecraft operated by the European Space Agency: Mars Express and INTEGRAL.

Mars Express (MEX), a long-lasting mission of the European Space Agency, has been exploring Mars since 2004. It is responsible for a wealth of scientific discoveries, including evidence of the presence of water (above and below the surface of the planet), an ample amount of three-dimensional renders of the surface, and a complete map of the chemical composition of Mars’s atmosphere. The scientific payload of MEX consists of seven instruments, which together with the onboard equipment have to be kept within their operating temperature ranges (from room temperature for some instruments, to temperatures as low as $-180^oC$ for others). In order to maintain these predefined operating temperatures, the spacecraft is equipped with an autonomous thermal system composed of 33 heater lines and coolers that consume a significant amount of the total generated electric power - leaving a fraction to be used for science operations. Therefore, given the age and the current condition of MEX, monitoring and optimally planning this consumption has a direct consequence on the longevity of the spacecraft and its mission \cite{lucas2017:mars,smc:2017,mex:mi,boumghar:spaceops,giros}.

INTEGRAL is a space observatory designed to monitor and detect gamma-rays with high sensitivity. Since its launch in 2002, it has been responsible for detecting iron quasars, investigating high energy gamma-ray burst as evidence of black-holes, supernovae remnants and active galactic nuclei (AGNs), as well as providing imaging and spectroscopic observations of astronomical events in both the $X$-ray range and optical wavelengths. During its 64-hour orbit around Earth (with an apogee of $\sim$140 000 km and perigee of $\sim$6 000 km), INTEGRAL passes through the Van Allen radiation belts, where radiation levels are high enough to potentially damage the onboard equipment. While the spacecraft is equipped with radiation sensors, these operate autonomously and are used for emergency instrument shutdowns, which are followed by lengthy recovery procedures. Accurately modeling and predicting the spacecraft’s position w.r.t these radiation belts is important. This allows for better control over activation/deactivation of onboard instruments and ultimately leading to optimal scientific output \cite{finn:integral}.

GalaxAI aims at addressing these tasks - by allowing for robust, accurate, and interpretable data analysis, it facilitates better and more informative mission planning. In the context of the MEX case study, this translates to better prediction of the thermal power consumption and therefore better estimation of the total available power, ultimately prolonging the mission itself. For INTEGRAL, on the other hand, better prediction of the Van Allen belts crossings would mean less recovery time and more time available for science. All of these predictive capabilities, coupled with visualisations and explainable model-outputs, are integrated into GalaxAI --  a versatile toolbox for better and informed decisions regarding the spacecraft's present and future status.

The paper is organized as follows. Section~\ref{sec:description} briefly describes the (software) architecture of GalaxAI. Next, Section~\ref{sec:pipelines} presents the machine learning pipelines available in GalaxAI. Section~\ref{sec:front} discusses the graphical user interface of the toolbox. Finally, Section~\ref{sec:conclusions} concludes the paper.
\section{Description of GalaxAI}\label{sec:description}

\begin{sidewaysfigure}
    \centering
    \includegraphics[width=\textwidth]{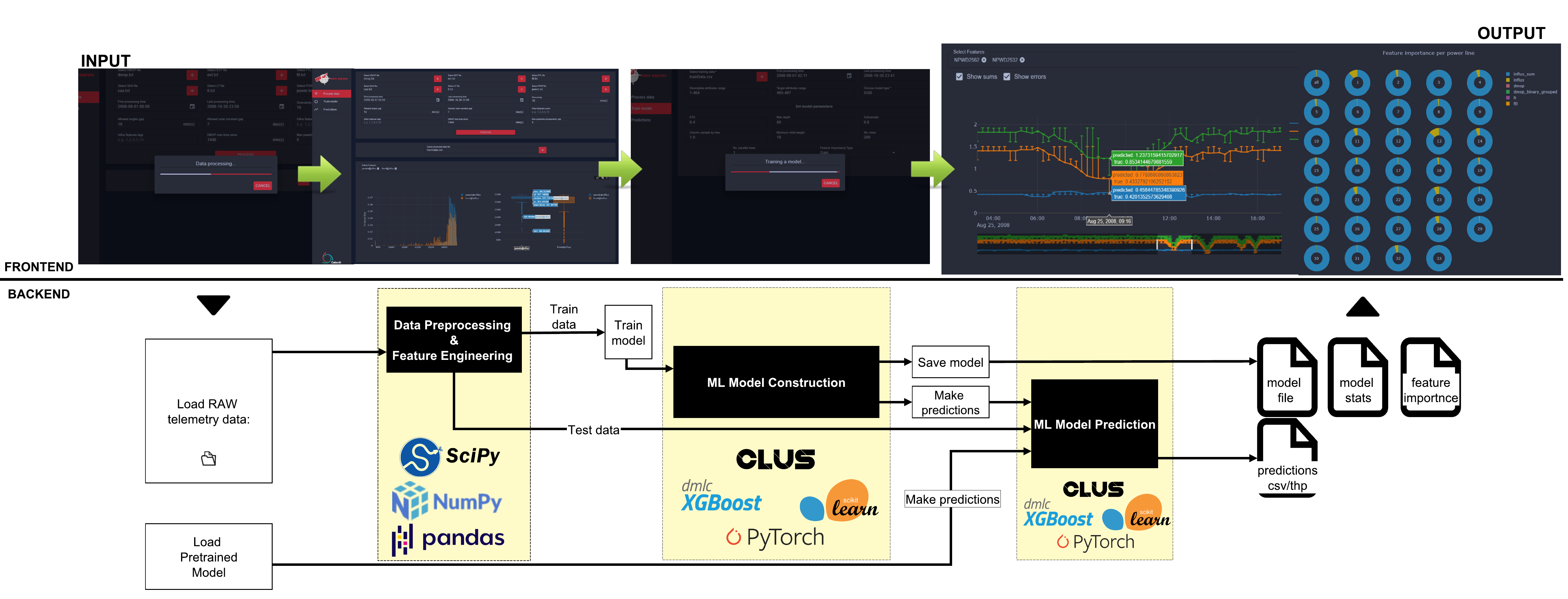}
    \caption{Overview of GalaxAI}
    \label{fig:overview}
\end{sidewaysfigure}

GalaxAI follows a two-layer design, consisting of a back-end and a front-end layer (Figure~\ref{fig:overview}). The cornerstone of GalaxaAI, its machine learning (ML) framework, is implemented as a part of the back-end layer. Besides modularity and easy maintenance, such implementation also allows certain data/compute intensive ML routines to be automated and executed on dedicated computing infrastructures. 

The ML framework consists of three major parts: (1) data preprocessing, feature engineering and selection, (2) model construction (learning), and (3) making predictions with a learned model. The first part includes various data preprocessing techniques and feature engineering algorithms designed and employed to pre-process the raw telemetry data pertaining to a particular spacecraft. 

The second part focuses on learning predictive models suitable for a considered data analysis task. More specifically, GalaxAI employs and supports various state-of-the-art machine-learning libraries including scikit-learn (for more 'traditional' machine learning methods)\cite{scikit-learn}, pytorch\cite{pytorch} (for deep neural networks methods) and CLUS\cite{clus} (for constructing predictive clustering trees). Depending on the task at hand, GalaxAI allows for the use of different suitable methods. For instance, in the case of predicting MEX's thermal power consumption, this relates to different methods for multi-output regression such as ensembles predicting clustering trees \cite{pcts}, gradient boosting ensembles\cite{xgb}, and fully-connected neural networks\cite{nn}. 

The third part of GalaxAI focuses on making predictions and visualising the findings. These range from simply plotting the predicted values to a more sophisticated analysis of the utility and relevance of the features used in the model construction phase. Moreover, GalaxAI also allows for running different simulation scenarios and ‘what-if’ analyses of spacecraft operations under different conditions and analysing the impact of specific sets of commands.


GalaxAI is also fully operable from a front-end layer through a Graphical User Interface (GUI). This enables users without any particular expertise in ML and software engineering to execute different parts-of or the complete data analysis pipeline. The interface employs React \cite{reactjs}, an open-source javascript framework used for front-end development of primarily web applications running on Node.js runtime environment \cite{nodejs}. This is utilized together with Electron \cite{electronjs}, an open-source platform for building desktop/offline applications. For visualizing the model outputs (i.e., predictions, feature importance diagrams, etc.), GalaxAI employs the interactive data visualization library Plotly.js \cite{plotly}.

\section{Machine Learning Pipelines}\label{sec:pipelines}


\newcommand{\gaiMEX}{GalaxAI-MEX}
\newcommand{\gaiINT}{GalaxAI-INTEGRAL}

We present two instantiations of GalaxAI, one for each of the spacecraft under consideration: \gaiMEX{} - focusing on analyses of the thermal power consumption of the Mars Express spacecraft and \gaiINT{} - for analyses of INTEGRAL's entry/exit times from the Van Allen belts. 

\subsection{\gaiMEX{}}

Recall that the back-end consists of three main parts that cover different stages of the pipeline. Here, we first briefly describe the input data as well as the three stages of the \gaiMEX{} pipeline.

\subsubsection{Input data}
GalaxAI-MEX processes six heterogeneous types of data. These include \textit{solar aspect angles (SAA)}, \textit{detailed mission operation plans (DMOP)}, \textit{flight dynamics timeline (FTL)}, \textit{various events (EVT)}, \textit{long-term data (LT)} and \textit{power data (PW)}. More specifically, SAA give the orientation of the spacecraft with respect to the Sun, while DMOP data give the commands that are issued with the spacecraft, together with the subsystem where each command is issued. FTL data give the pointing-events (towards Mars, the Earth, etc.) and EVT lists various MEX-orbit-related events, such as entering/exiting umbra, passing through the extreme points (apo- and pericenter) of the orbit, etc. Finally, LT data contain the values of some physical quantities that can be computed far into the future (e.g., the distance between Mars and the Sun) and PW data gives the electrical currents through each of the 33 electrical heaters onboard the spacecraft.

\subsubsection{Data preprocessing} 
Given the heterogeneous raw data, the preprocessing within \gaiMEX{} includes data alignment, feature construction, aggregation of the power data, and data cleansing. The data is first aligned to a given time-granularity (e.g., one entry per 15 minutes), as the entries from various data files are time-stamped, but are recorded at different (and irregular) paces. The next step of \textit{feature construction} creates features (often by joining different data sources) used for learning the predictive models. This step is necessary since the data in its raw format is not directly usable by a machine learning algorithm. The \textit{aggregation of the power-data} includes computation of the average electrical current (e.g., for every 15 minutes) for each of the thermal consumers. Finally, \textit{data cleansing} removes/imputes records with missing values from the data. Details describing the feature construction procedure(s) are given by \cite{mex:mi}. At the end of this stage, the constructed data set can be readily used for model learning. In addition to the data set itself, a metafile is created that contains all the information necessary for reproducing the learning experiments.

\subsubsection{Learning models}
The constructed data set from the previous stage is used here for constructing a predictive model. To this end, \gaiMEX{} implements several different machine learning methods. Namely, it implements unified wrappers for XGBoost \cite{xgb}, PCT-based ensembles \cite{pcts}, (deep) fully-connected neural networks \cite{nn}, as well as for all models implemented in the scikit-learn toolbox \cite{scikit-learn}. Moreover, it implements feature ranking methods to provide a better understanding of the models and the predictions.  In particular, it provides three feature ranking scores \cite{petkomat:mtr,petkovic2020fuji} that calculate the feature importance/relevance: (1) random forest mechanism, (2) GENIE3, and (3) Symbolic scores. The first score can be applied to arbitrary types of machine learning models (since it is based on random permutations of the values, and estimating the difference in the errors caused by it). The GENIE3 and Symbolic scores can be calculated only for tree-based ensemble models. All three scores can be calculated for subsets of data examples selected by the user.

The default values of the hyperparameters for most of the machine learning methods were selected in a comprehensive experimental study that used the MEX data between 2008 and 2020. Once a predictive model has been learned it is ready for use in making predictions for unseen (and/or future) data. The learned model is stored together with an additional meta file (containing all the parameters used for learning the model), allowing for shared analysis with other users as well as reproducible analysis. For an extensive study on the performance of these models in the context of predicting MEX's thermal power consumption, we refer the reader to \cite{mex:mi,giros}.

\subsubsection{Making predictions} 
At the final stage, the constructed predictive models are employed for making predictions. \gaiMEX{} employs various evaluation strategies for estimating the performance of the learned models and the quality of the predictions. Moreover, it includes a mechanism for interpreting model outputs in terms of feature importance diagrams, that depict the influence of the input features on the resulting predictions. These evaluation statistics together with the model predictions are the output of \gaiMEX{}.

\subsection{\gaiINT{}}
Similarly as before, here we first describe the raw data used in the pipeline, its preprocessing, representation, and preparation for use by machine learning algorithms. Then, we describe the implemented machine learning methods and their use to analyze INTEGRAL data.

\subsubsection{Input data}
To determine when the spacecraft enters the Van Allen belts, we rely on the onboard IREM measurements, taken every 8 seconds. Since they can be very noisy, we take median values from bins with a coarser time granularity (5-15 minutes). The times of entries into and exits from the belt are determined through thresholding these IREM measurements. More specifically, when the counts are above 600 electron counts per second, the spacecraft is determined to be inside the belts. 

The orbit of each revolution of the spacecraft is defined by 12 orbital elements that include (1) perigee time, (2) perigee altitude, (3) apogee time, (4) apogee altitude, (5) longitude, above which perigee is located, (6) semi-major axis, (7) eccentricity, (8) inclination, (9) right ascension of the ascending node, (10) argument of perigee, (11) period, and (12) period difference from the previous revolution. We also take into account the eclipse times, when the spacecraft is shadowed from the Sun by the Earth or the Moon. The orbital elements and eclipse times are available for several months into the future and form the basis from which we engineer features that the models use to predict entry/exit times.

\begin{figure*}[!t]
     \centering
     \begin{subfigure}[b]{0.49\textwidth}
         \centering
         \includegraphics[width=\textwidth]{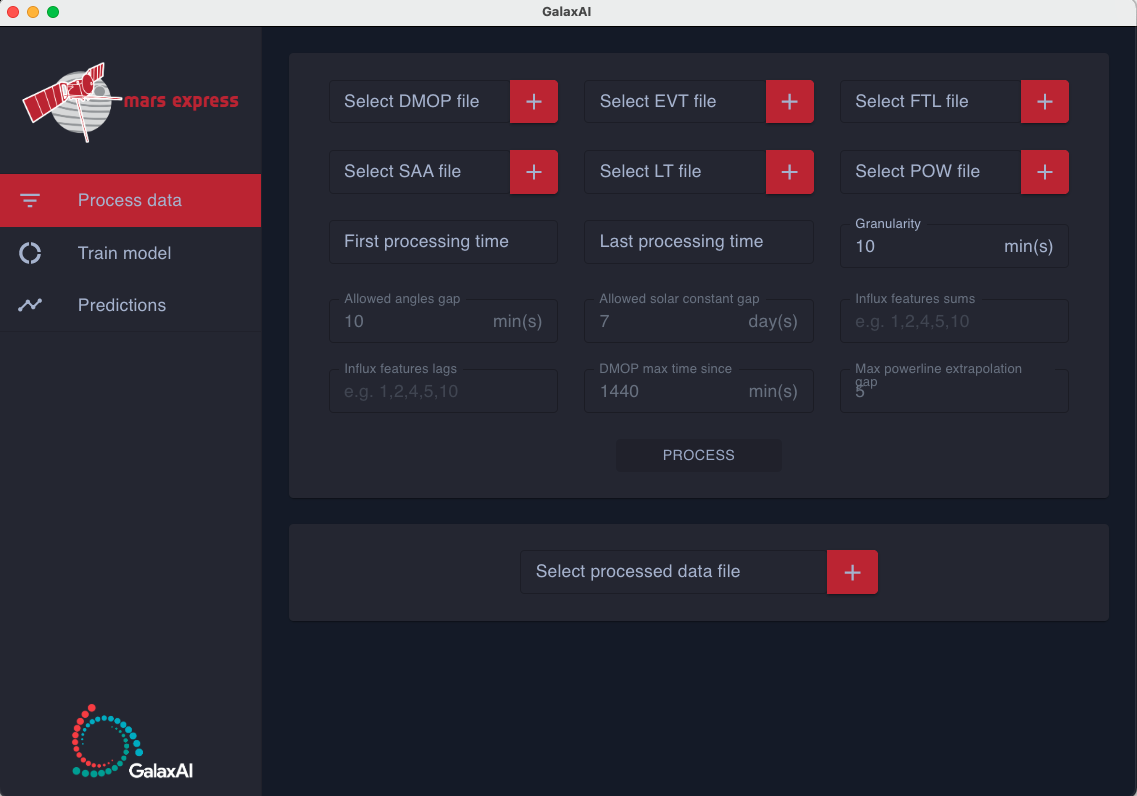}
     \end{subfigure}
     \begin{subfigure}[b]{0.49\textwidth}
         \centering
         \includegraphics[width=\textwidth]{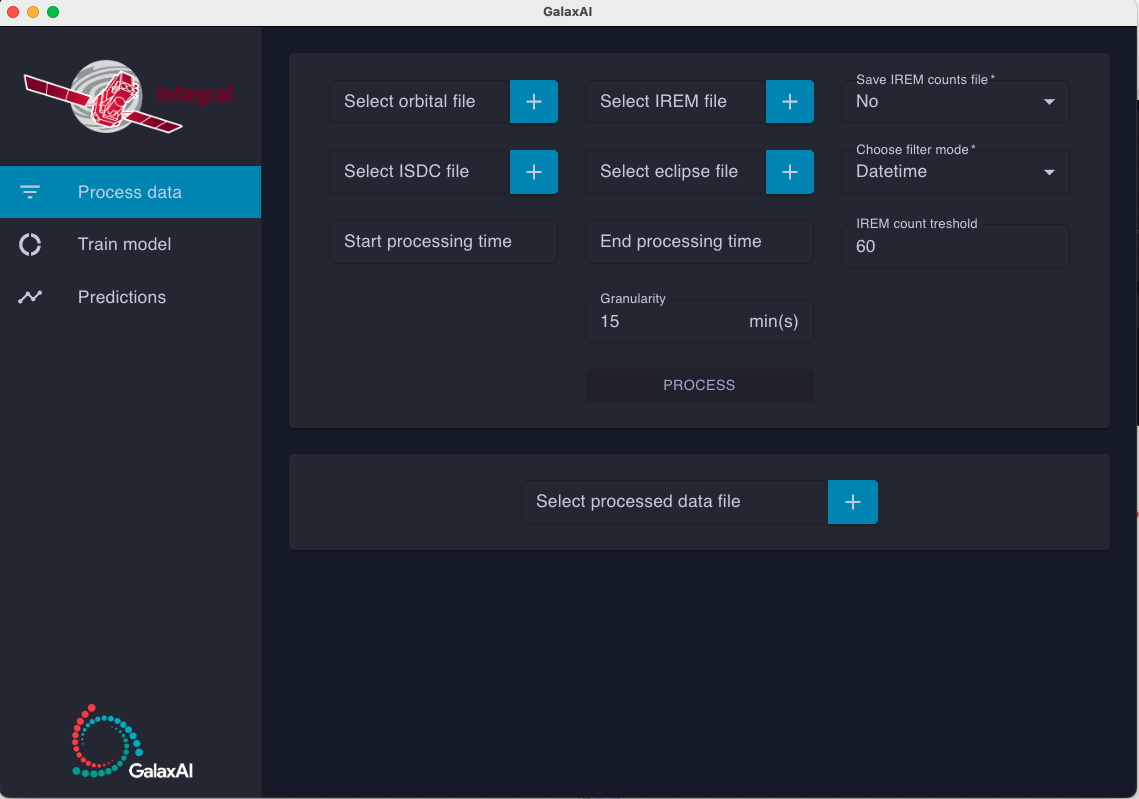}
     \end{subfigure}
        \caption{GUI of \gaiMEX{}(left-hand side) and \gaiINT{} (right-hand side) GUIs.}
        \label{fig:gui}
\end{figure*}

\subsubsection{Data representation and preprocessing}
Since we are interested in the orbital position of the spacecraft, all timestamps are first transformed to \emph{phase} values relative to the current revolution. The phase values range from 0 at perigee to 1 at the next perigee. For this task, we consider two data representations -- \emph{positional} and \emph{per-revolution}. The former, positional representation, is similar to the one proposed by Finn et al.~\cite{finn:integral}. Here, the data is ordered in a series where examples describe the state of the spacecraft using the orbital elements and the IREM counts (or binary indicators whether INTEGRAL is in the belts or not). Thus one can consider two different tasks: regression (when predicting the IREM counts) or classification (when predicting the binary indicator).



In the \emph{per-revolution} representation, each example describes one revolution of the spacecraft using the 12 orbital elements, together with the times when the spacecraft enters and exits Earth's umbra and penumbra and the penumbra of the Moon. If any of these events do not occur for a given revolution, we use the apogee time as a fill-in value because apogee is furthest removed from the events of interest (belts entry/exit) which happen near perigee. This yields a total of 18 features.In this representation, we can directly predict the entry/exit times (or altitudes) for a given revolution. This gives us two target variables (predictands) and we can treat the problem as a multi-target regression task.

Each revolution takes approximately 64 hours. In the positional representation with a 15-minute granularity, there are 264 time\-stamps (examples) during each revolution, which is a single example in the per-revolution representation. The per-revolution representation is therefore much more compact. By performing a comprehensive set of experiments, we found that it provides better predictions for entry and exit times.


\begin{figure*}[!b]
    \centering
    \includegraphics[width=\textwidth]{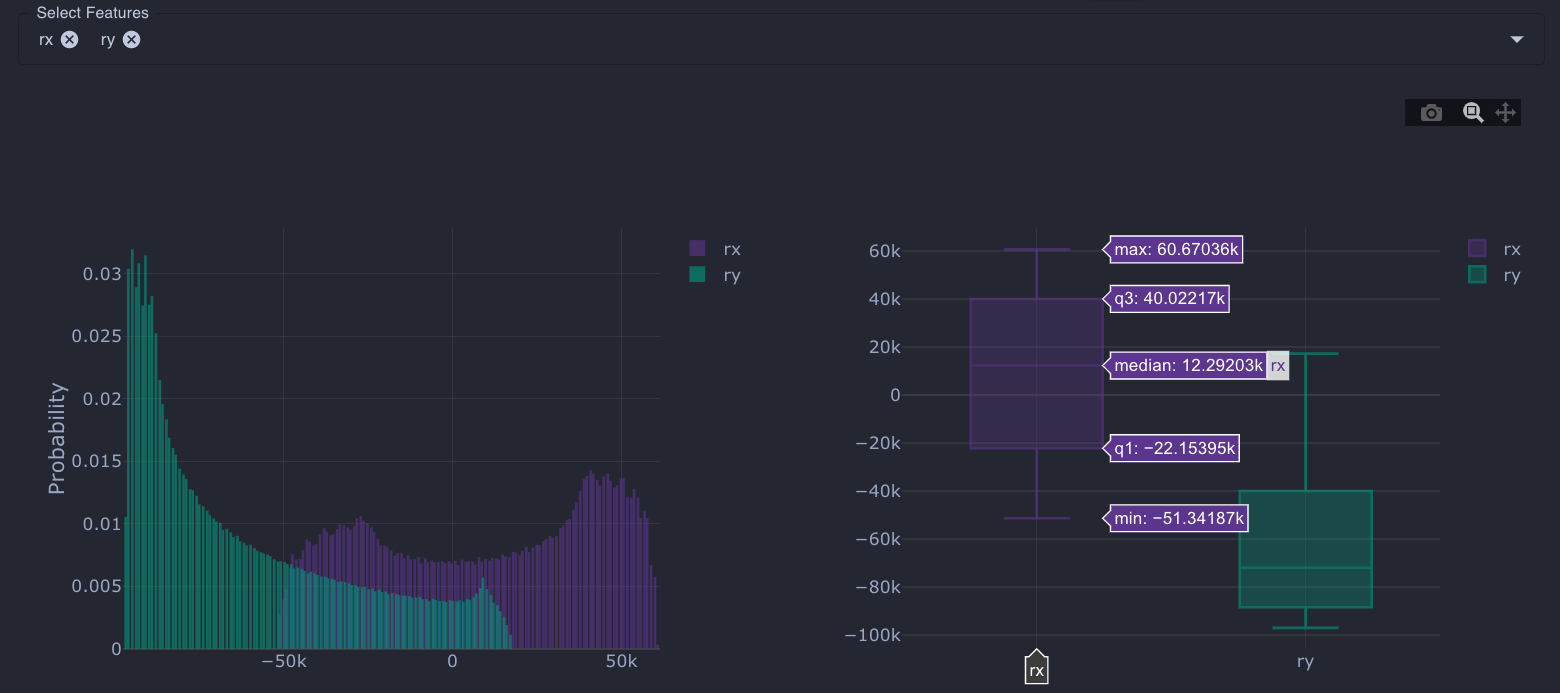}
    \caption{\gaiINT{}: Exploratory data analysis plots.}
    \label{fig:EDA}
\end{figure*}

\subsubsection{Learning models}
Similarly to \gaiMEX{}, \gaiINT{} implements several machine learning methods: 
(1) $k$-nearest neighbor regressor (KNN), (2) random forest ensembles of regression trees (RF), (3) extreme gradient boosting ensembles of regression trees (XGB), (4) gradient boosting ensembles of regression trees with quantile loss (GB), (5) fully connected neural networks (FCNN), (6) recurrent neural networks (RNN) with gated recurrent units. In particular, for some of the methods (KNN, RF, and GB), \gaiINT{} employs the \emph{scikit-learn} \cite{scikit-learn} implementations. For XGB, \gaiINT{} uses the \emph{xgboost} Python library \cite{xgb}. \gaiINT{} implements both XGB and GB since GB supports quantile regression. This allows models to predict the conditional median instead of the mean, which can help deal with noisy data. 

The neural network models are implemented in the Pytorch framework \cite{pytorch}. While RNNs, in particular, are well suited for time-series data by design, the remaining methods require additional engineering for taking the temporal aspect into account. To this end, we add additional historical information to each example, i.e., each example has access to the features of the previous $n$ examples and the targets of the previous $m$ examples. We refer to these values as \emph{feature history} and \emph{autoregression history}, respectively.

An important consideration for the methods is the number of target variables: in the positional representation, there is only \emph{one} target (the IREM count rate or in-belt indicator), while in the per-revolution representation, there are \emph{two} targets---the entry and exit altitudes/phases. Most of the methods used can handle predicting two target variables with a single model (global approach). The exceptions are XGB and GB, where \gaiINT{} constructs a separate model for each target variable (local approach).

Prior to model learning, the data are standardized, but nevertheless, the model predictions, in the end, are inversely transformed to get values on the original scale. In terms of handling missing values, in the learning phase, examples with missing targets are excluded from the learning set. In turn, when evaluating a model, the prediction errors are only calculated on non-missing values. When using autoregression history, missing target values can result in missing feature values as well. In such cases, the missing feature values are imputed with the mean value for that variable in the learning set.


\begin{figure*}[!t]
    \centering
    \includegraphics[width=\textwidth]{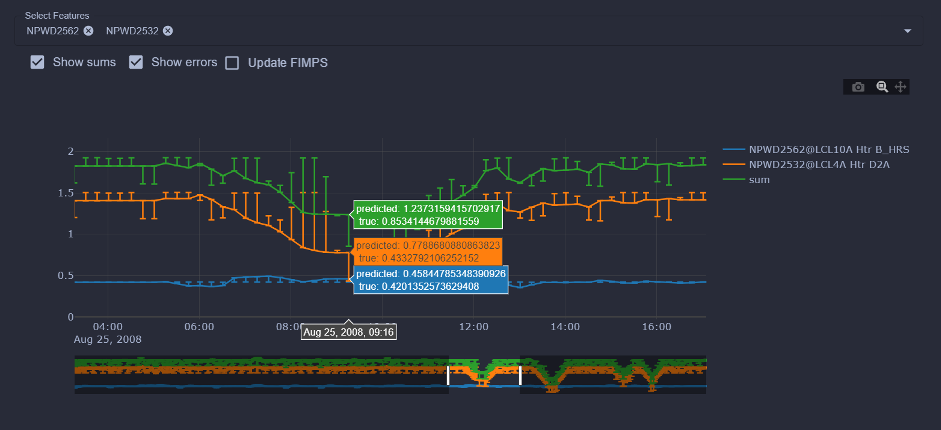}
    \caption{\gaiMEX{}: Scatterplot visualizing the model predictions for NPWD2562 and NPWD2532 for 25.08.2008}
    \label{fig:predictions}
\end{figure*}

\section{Interaction with data and models}\label{sec:front}

The machine learning pipelines that are executable through the GalaxAI toolbox are well-structured, documented, and accessible by a command-line interface, albeit mostly well suited for data science practitioners. Nevertheless, such usage scenarios can create some serious non-trivial challenges when used by engineers and operators who do not have prior experience working with ML-based frameworks. Such challenges include the choice of the predictive model, choosing and setting model parameters, interpretability of the model as well as explainability of its findings. The latter two, in particular, are very important when it comes to increasing the trustworthiness and facilitating the utility of predictive models in practice, especially when working with black-box models such as neural networks.

\begin{figure}[!b]
    \centering
    \includegraphics[width=\linewidth]{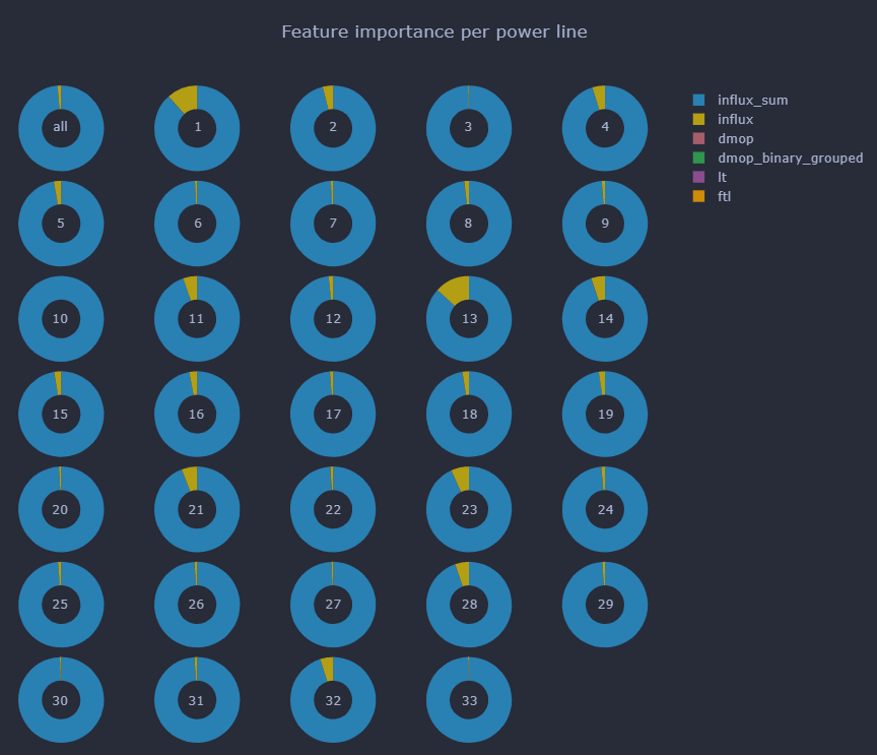}
    \caption{\gaiMEX{}: Doughnut charts visualizing the importance of the descriptive features (per feature category).}
    \label{fig:doughnut}
\end{figure}

\begin{figure*}[!t]
    \centering
    \includegraphics[width=0.9\textwidth]{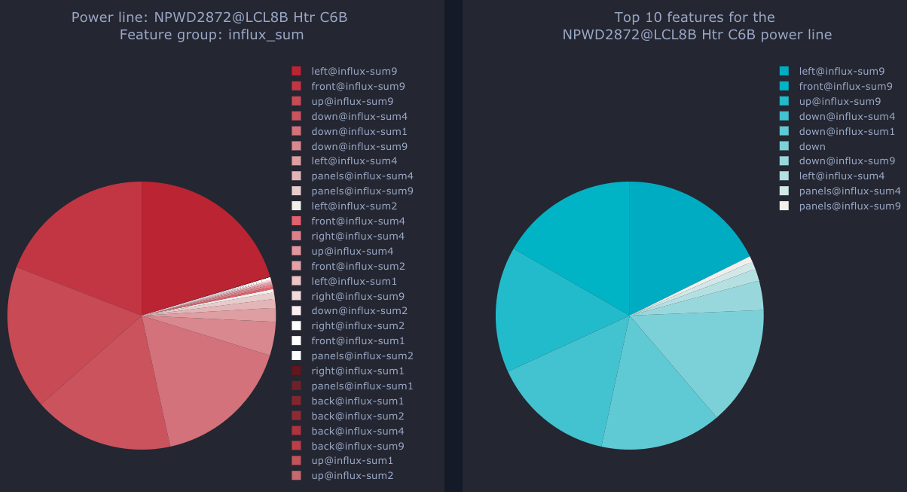}
    \caption{\gaiMEX{} : Charts visualizing the importance of the selected feature groups for NPWD2872.}
    \label{fig:pie}
\end{figure*}

GalaxAI addresses these challenges by employing a user-friendly graphical user interface (GUI) for executing ML pipeline(s), allowing for both visual exploratory data analysis and visualization of the model results. In particular, GalaxAI facilitates seamless execution of the ML pipelines by providing pre-selected learning methods with optimal parameters (selected based on comprehensive experimental study). Next, the interactive nature of the visualizations enables the domain experts to perform exploratory data analysis on the preprocessed data and interpret the obtained models and predictions. Moreover, GalaxAI allows for performing various `what-if' analysis scenarios by excluding data examples and/or features. The `what-if' analysis provides additional means to integrate existing expert knowledge into the learning process thus further improving the results. 

Figure~\ref{fig:gui} depicts the initial viewscreen of GalaxAI, with respect to \gaiMEX{} and \gaiINT{}. The users can perform several tasks related to the data analysis pipeline: (i) pre-process and analyse data (ii) use pre-trained models to perform predictions and analyse the results and (iii) execute the complete end-to-end pipeline from loading raw data to learning a predictive model, to analysing the results. For executing each task, the user needs to only provide input data/model required for the specific analysis scenario. For easier usage, most of the parameters are already set by default to their optimal value, but the users can always change them and explore other parameters based on their experience with the domain and the data.

\subsubsection{Exploratory Data Analysis}

 Within GalaxAI, we have implemented different interactive diagrams (histograms and boxplots) that enable users to explore the data. More specifically, the diagrams allow users to select time ranges for visualization as well as to select several variables at the same time. Figure~\ref{fig:EDA} depicts these diagrams as implemented in \gaiINT{} for selected features’ distributions (similar diagrams are also available in \gaiMEX{}).



\subsubsection{Predictions Visualization}

In terms of visualising the model output, GalaxAI implements several diagrams pertaining to visualization of the obtained predictions and visualization of the influence/importance of the descriptive features. The former involves an interactive scatterplot for visual inspection of the predicted values. Figure~\ref{fig:predictions} shows such a diagram as implemented within \gaiMEX{}, for visualizing the predicted thermal power consumption of MEX. The diagrams can visualise the predictions of multiple thermal powerlines simultaneously or plot their cumulative predicted value at each time point. The latter gives a more general overview, allowing for a quick assessment of the predictive analysis. Moreover, in scenarios when the test data set, provided at input, contains the true values of the predictands, both the true and predicted values will be displayed coupled with error bars at each time point depicting their discrepancy. Similar diagrams are available also within \gaiINT{}.

\subsubsection{Feature Importance Visualization}

GalaxAI allows for visual inspection of the feature influence within the predictive models. More specifically, it implements a special type of interactive 'doughnut' charts (see Figure~\ref{fig:doughnut}) and pie charts (see Figure~\ref{fig:pie}) for global and local visualization of the importance of the predictive features (calculated as three scores) to the predictive task at hand. These charts provide the means for better explainability of both models and predictions. Namely, a feature is important when a model relies on it for predictions. Thus, by observing the importance of each feature used for learning a model, one can explain, to a certain extent, the model’s predictions.


In the context of \gaiMEX{}, the visual feature-importance analysis is performed with 34 doughnut charts (Figure \ref{fig:doughnut}) - one for each of the 33 power lines and one that aggregates the feature importance across all power lines. These plots provide global, and compressed, insight into the model's behavior with respect to different feature categories. Alternatively, the user can select each doughnut (corresponding to a powerline) which will result in plotting the 10-most important features (across all feature groups) for predicting that particular powerline (Figure \ref{fig:pie}). For a more detailed look at individual feature importance diagrams, the user can also select a feature category that will render them in a pie chart (Figure \ref{fig:pie}). Analogously, these charts are also available in \gaiINT{}, with the detailed feature visualisation available by default. Finally, GalaxAI supports interactive updates of these charts with respect to a selected time-interval of interest (by controlling the slider on the bottom of Figure~\ref{fig:predictions}), thus providing even more insight into the behaviour of the predictive model and the resulting prediction.

\section{Conclusions}\label{sec:conclusions}

Spacecraft monitoring and operation involve many challenging tasks and decisions -  most often based on analysis of large volumes of complex, multimodal, and heterogeneous telemetry data. These analyses, in turn, are used for monitoring the spacecraft's health as well as short/long-term operations planning. Therefore they need to be very accurate, but more importantly, they need to provide a better understanding of the spacecraft's status and support the decisions of the mission operators and engineers.


In this work, we present GalaxAI - a versatile machine learning toolbox for accurate, efficient, and interpretable end-to-end data analysis of spacecraft telemetry data. It implements various machine learning pipelines that are well-structured, documented, and accessible by command-line interface useful for data science practitioners. It also offers a user-friendly graphical interface for executing the underlying machine learning pipelines and performing visual exploratory data analysis and model visualizations.

We show the utility GalaxAI on two use-cases from two spacecraft: i) analysis and planning of Mars Express thermal power consummation and ii) predictive analysis of INTEGRAL's crossing through the Van Allen belts. The interactive nature of the visualizations enables the domain experts to perform exploratory data analysis and interpret the obtained models and predictions. Moreover, GalaxAI allows for performing various `what-if' analyses thus providing means to integrate existing expert knowledge into the learning process and, in turn, to further enhance it. While in this work, we showcase two use-cases, GalaxAI is general, modular, and easily extensible to other data analysis tasks, missions, and spacecraft.








\section*{Acknowledgment}

The authors acknowledge the financial support of ESA through the project GalaxAI: Machine learning for space operations. Also, AK, MP, TS, PP, NS and DK acknowledge the support of the Slovenian Research Agency through the research program No.~P2-0103 and research project No.~J2-9230. 

\bibliographystyle{IEEEtran}
\bibliography{references}

\begin{thebibliography}{10}
\providecommand{\url}[1]{#1}
\csname url@samestyle\endcsname
\providecommand{\newblock}{\relax}
\providecommand{\bibinfo}[2]{#2}
\providecommand{\BIBentrySTDinterwordspacing}{\spaceskip=0pt\relax}
\providecommand{\BIBentryALTinterwordstretchfactor}{4}
\providecommand{\BIBentryALTinterwordspacing}{\spaceskip=\fontdimen2\font plus
\BIBentryALTinterwordstretchfactor\fontdimen3\font minus
  \fontdimen4\font\relax}
\providecommand{\BIBforeignlanguage}[2]{{%
\expandafter\ifx\csname l@#1\endcsname\relax
\typeout{** WARNING: IEEEtran.bst: No hyphenation pattern has been}%
\typeout{** loaded for the language `#1'. Using the pattern for}%
\typeout{** the default language instead.}%
\else
\language=\csname l@#1\endcsname
\fi
#2}}
\providecommand{\BIBdecl}{\relax}
\BIBdecl

\bibitem{McGovern2011}
A.~McGovern and K.~L. Wagstaff, ``Machine learning in space: extending our
  reach,'' \emph{Machine Learning}, vol.~84, no.~3, pp. 335--340, 2011.

\bibitem{Yairi17:jrnl}
T.~Yairi, N.~Takeishi, T.~Oda, Y.~Nakajima, N.~Nishimura, and N.~Takata, ``A
  data-driven health monitoring method for satellite housekeeping data based on
  probabilistic clustering and dimensionality reduction,'' \emph{IEEE
  Transactions on Aerospace and Electronic Systems}, vol.~53, no.~3, pp.
  1384--1401, 2017.

\bibitem{Carlton2018:jrnl}
A.~Carlton, R.~Morgan, W.~Lohmeyer, and K.~Cahoy, ``Telemetry fault-detection
  algorithms: Applications for spacecraft monitoring and space environment
  sensing,'' \emph{Journal of Aerospace Information Systems}, vol.~15, no.~5,
  pp. 239--252, 2018.

\bibitem{Ahn2020:jrnl}
H.~Ahn, D.~Jung, and H.-L. Choi, ``Deep generative models-based anomaly
  detection for spacecraft control systems,'' \emph{Sensors}, vol.~20, no.~7,
  pp. 1991:1--20, 2020.

\bibitem{Wang2019:jrnl}
C.~Wang, N.~Lu, Y.~Cheng, and B.~Jiang, ``A telemetry data based diagnostic
  health monitoring strategy for in-orbit spacecrafts with component
  degradation,'' \emph{Advances in Mechanical Engineering}, vol.~11, no.~4, pp.
  1--14, 2019.

\bibitem{heras2014:jrnl}
J.-A. Martínez-Heras and A.~Donati, ``Enhanced telemetry monitoring with
  novelty detection,'' \emph{AI Magazine}, vol.~35, no.~4, pp. 37--46, 2014.

\bibitem{Ibrahim19:jrnl}
S.~K. Ibrahim, A.~Ahmed, M.~A.~E. Zeidan, and I.~E. Ziedan, ``Machine learning
  methods for spacecraft telemetry mining,'' \emph{IEEE Transactions on
  Aerospace and Electronic Systems}, vol.~55, no.~4, pp. 1816--1827, 2019.

\bibitem{pilastre2020}
B.~Pilastre, L.~Boussouf, S.~D’Escrivan, and J.-Y. Tourneret, ``Anomaly
  detection in mixed telemetry data using a sparse representation and
  dictionary learning,'' \emph{Signal Processing}, vol. 168, p. 107320, 2020.

\bibitem{lucas2017:mars}
L.~Lucas and R.~Boumghar, ``Machine learning for spacecraft operations support
  - {The Mars Express Power Challenge},'' in \emph{Proceedings of the Sixth
  International Conference on Space Mission Challenges for Information
  Technology, {SMC-IT 2017}}, 2017, pp. 82--87.

\bibitem{smc:2017}
M.~Breskvar, D.~Kocev, J.~Levati\'{c}, A.~Osojnik, M.~Petkovi\'{c},
  N.~Simidjievski, B.~\v{Z}enko, R.~Boumghar, and L.~Lucas, ``Predicting
  thermal power consumption of the mars express satellite with machine
  learning,'' in \emph{Proceedings of the {6th International Conference on
  Space Mission Challenges for Information Technology (SMC-IT)}}, 2017, pp.
  88--93.

\bibitem{mex:mi}
M.~Petkovi\'{c}, R.~Boumghar, M.~Breskvar, S.~D\v{z}eroski, D.~Kocev,
  J.~Levati\'{c}, L.~Lucas, A.~Osojnik, B.~\v{Z}enko, and N.~Simidjievski,
  ``Machine learning for predicting thermal power consumption of the mars
  express spacecraft,'' \emph{IEEE Aerospace and Electronic Systems Magazine},
  vol.~34, no.~7, pp. 46--60, 2019.

\bibitem{boumghar:spaceops}
R.~Boumghar, L.~Lucas, and A.~Donati, ``Machine learning in operations for the
  mars express orbiter,'' in \emph{15th International Conference on Space
  Operations}, Marseille, France, 2018.

\bibitem{giros}
M.~Petkovi\'{c}, L.~Lucas, D.~Kocev, S.~D\v{z}eroski, R.~Boumghar, and
  N.~Simidjievski, ``Quantifying the effects of gyroless flying of the mars
  express spacecraft with machine learning,'' in \emph{Proceedings of the {2019
  IEEE International Conference on Space Mission Challenges for Information
  Technology (SMC-IT)}}, 2019, pp. 9--16.

\bibitem{finn:integral}
T.~Finn, R.~Boumghar, J.~Martínez-Heras, and A.~Georgiadou, ``Machine learning
  modeling methods for radiation belts profile predictions,'' in
  \emph{{Proceedings of the 2018 SpaceOps Conference}}.\hskip 1em plus 0.5em
  minus 0.4em\relax American Institute of Aeronautics and Astronautics, Inc.,
  2018, pp. 1--11.

\bibitem{scikit-learn}
F.~Pedregosa, G.~Varoquaux, A.~Gramfort, V.~Michel, B.~Thirion, O.~Grisel,
  M.~Blondel, P.~Prettenhofer, R.~Weiss, V.~Dubourg, J.~Vanderplas, A.~Passos,
  D.~Cournapeau, M.~Brucher, M.~Perrot, and E.~Duchesnay, ``Scikit-learn:
  Machine learning in {P}ython,'' \emph{Journal of Machine Learning Research},
  vol.~12, pp. 2825--2830, 2011.

\bibitem{pytorch}
A.~Paszke, S.~Gross, F.~Massa, A.~Lerer, J.~Bradbury, G.~Chanan, and T.~e.~a.
  Killeen, ``Pytorch: An imperative style, high-performance deep learning
  library,'' in \emph{Advances in Neural Information Processing Systems
  32}.\hskip 1em plus 0.5em minus 0.4em\relax Curran Associates, Inc., 2019,
  pp. 8024--8035.

\bibitem{clus}
``\textsc{Clus}: Predictive clustering toolbox,''
  \url{http://source.ijs.si/ktclus/clus-public}, accessed: 2021-02-22.

\bibitem{pcts}
\BIBentryALTinterwordspacing
D.~Kocev, C.~Vens, J.~Struyf, and S.~Džeroski, ``Tree ensembles for predicting
  structured outputs,'' \emph{Pattern Recognition}, vol.~46, no.~3, pp. 817 --
  833, 2013. [Online]. Available:
  \url{http://www.sciencedirect.com/science/article/pii/S003132031200430X}
\BIBentrySTDinterwordspacing

\bibitem{xgb}
\BIBentryALTinterwordspacing
T.~Chen and C.~Guestrin, ``{XGBoost}: A scalable tree boosting system,'' in
  \emph{Proceedings of the 22nd ACM SIGKDD International Conference on
  Knowledge Discovery and Data Mining}, ser. KDD '16.\hskip 1em plus 0.5em
  minus 0.4em\relax New York, NY, USA: ACM, 2016, pp. 785--794. [Online].
  Available: \url{http://doi.acm.org/10.1145/2939672.2939785}
\BIBentrySTDinterwordspacing

\bibitem{nn}
I.~Goodfellow, Y.~Bengio, and A.~Courville, \emph{Deep Learning}.\hskip 1em
  plus 0.5em minus 0.4em\relax The MIT Press, 2016.

\bibitem{reactjs}
{Facebook Inc.}, ``React,'' \url{https://reactjs.org/}, Menlo Park, CA, USA,
  2021.

\bibitem{nodejs}
{OpenJS Foundation}, ``Node.js,'' \url{https://nodejs.org/}, San Francisco, CA,
  USA, 2021.

\bibitem{electronjs}
------, ``Electron.js,'' \url{https://www.electronjs.org/}, San Francisco, CA,
  USA, 2021.

\bibitem{plotly}
{Plotly Technologies Inc.}, ``Collaborative data science,''
  \url{https://plot.ly}, Montreal, QC, 2015.

\bibitem{petkomat:mtr}
M.~Petkovi\'{c}, D.~Kocev, and S.~D\v{z}eroski, ``Feature ranking for
  multi-target regression,'' \emph{Machine Learning}, vol. 109, pp. 1179--1204,
  2020.

\bibitem{petkovic2020fuji}
M.~Petković, B.~Škrlj, D.~Kocev, and N.~Simidjievski, ``Fuzzy jaccard index:
  A robust comparison of ordered lists,'' \emph{CoRR}, vol. abs/2008.02216,
  2022.

\end{thebibliography}

\end{document}